\pdfoutput=1%
\documentclass[runningheads,a4paper]{llncs}

\usepackage{times}
\usepackage{amssymb}
\usepackage{graphicx}
\usepackage{url}
\usepackage[normalem]{ulem}

\usepackage{multirow}
\usepackage{pifont}
\usepackage{xcolor}
\usepackage{hyperref}
\usepackage{enumitem}

\newcommand{\YES}{\ding{51}}
\newcommand{\NO}{{\color{gray!50}\ding{55}}}
\newenvironment{citemize}{\begin{itemize}[noitemsep,topsep=0pt]}{\end{itemize}}

\newcommand{\keywords}[1]{\par\addvspace\baselineskip
\noindent\keywordname\enspace\ignorespaces#1}

\urldef{\mailsa}\path|{straka, strakova, hajic}@ufal.mff.cuni.cz|

\begin{document}

\title{Czech Text Processing with Contextual Embeddings:
\\POS Tagging, Lemmatization, Parsing and NER}

\titlerunning{Czech Text Processing with Contextual Embeddings}

\author{Milan Straka \and Jana Strakov\'{a} \and Jan Haji\v{c}}


\authorrunning{Milan Straka et al.}

\institute{Charles University, Faculty of Mathematics and Physics,\\Institute of Formal and Applied Linguistics \\
\url{http://ufal.mff.cuni.cz} \\
\mailsa
}

\index{Straka, Milan}
\index{Strakov\'{a}, Jana}
\index{Haji\v{c}, Jan}

\toctitle{} \tocauthor{}

\maketitle

%
%
%
%
\begin{abstract}
Contextualized embeddings, which capture appropriate word meaning depending on context, have recently been proposed. We evaluate two methods for precomputing such embeddings, BERT and Flair, on four Czech text processing tasks: part-of-speech (POS) tagging, lemmatization, dependency parsing and named entity recognition (NER). The first three tasks, POS tagging, lemmatization and dependency parsing, are evaluated on two corpora: the Prague Dependency Treebank 3.5 and the Universal Dependencies 2.3. The named entity recognition (NER) is evaluated on the Czech Named Entity Corpus 1.1 and 2.0. We report state-of-the-art results for the above mentioned tasks and corpora.
\keywords{contextualized embeddings, BERT, Flair, POS tagging, lemmatization, dependency parsing, named entity recognition, Czech}
\end{abstract}

\section{Introduction}

Recently, a novel way of computing word embeddings has been proposed. Instead of computing one word embedding for each word which sums over all its occurrences, ignoring the appropriate word meaning in various contexts, the \textit{contextualized embeddings} are computed for each word occurrence, taking into account the whole sentence. Three ways of computing such contextualized embeddings have been proposed: ELMo \cite{Peters2018}, BERT \cite{BERT} and Flair \cite{Akbik}, along with precomputed models.

Peters et al. (2018) \cite{Peters2018} obtain the proposed embeddings, called \textit{ELMo}, from internal states of deep bidirectional language model, pretrained on a large corpus. Akbik et al. (2018) \cite{Akbik} introduced \textit{Flair}, contextualized word embeddings obtained from internal states of a character-level bidirectional language model, thus significantly increasing state of the art of POS tagging, chunking and NER tasks. Last, but not least, Devlin et al. (2018) \cite{BERT} employ a Transformer \cite{vaswani:2017} to compute contextualized embeddings from preceeding and following context at the same time, at the cost of increased processing costs. The new \textit{BERT} embeddings achieved state-of-the-art results in eleven natural language tasks.

Using two of these methods, for which precomputed models for Czech are available, namely BERT and Flair, we present our models for four NLP tasks: part-of-speech (POS) tagging, lemmatization, dependency parsing and named entity recognition (NER). Adding the contextualized embeddings as optional inputs in strong artificial neural network baselines, we report state-of-the-art results in these four tasks.

\section{Related Work}

As for the Prague Dependency Treebank (PDT) \cite{PDT3.5}, most of the previous works are non-neural systems with one exception of \cite{EMNLP2018} who hold the state of the art for Czech POS tagging and lemmatization, achieved with the recurrent neural network (RNN) using end-to-end trainable word embeddings and character-level word embeddings. Otherwise, Spoustov\'{a} et al. (2009) \cite{spoustova09} used an averaged perceptron for POS tagging. For parsing the PDT, Holan and {Z}abokrtsk\'{y} (2006) \cite{holan:zabokrtsky:2006} and Nov\'{a}k and \v{Z}abokrtsk\'{y} (2007) \cite{novak:zabokrtsky:2007} used a combination of non-neural parsing techniques .

In the multilingual shared task \textit{CoNLL 2018 Shared Task: Multilingual
Parsing from Raw Text to Universal Dependencies} \cite{CoNLL2018}, raw text is processed and the POS tagging, lemmatization and dependency parsing are evaluated on the Universal Dependencies (UD) \cite{UD}. Czech is one of the $57$ evaluated languages. Interestingly, all $26$ participant systems employed the artificial neural networks in some way. Of these, $3$ participant systems 
used (a slightly modified variant of) the only newly presented contextualized embeddings called ELMo \cite{Peters2018}, most notably one of the shared task winners~\cite{HIT-SCIR}. BERT and Flair were not available at the time.

For the Czech NER, Strakov\'{a} et al. (2016) \cite{Strakova2016} use an artificial neural network with word- and character-level word embeddings to perform NER on the Czech Named Entity Corpus (CNEC) \cite{Sevcikova2007,CNEC1.1-data,CNEC2.0-data}.

\section{Datasets}
\label{section:datasets}

\subsubsection{Prague Dependency Treebank 3.5}

The \textit{Prague Dependency Treebank 3.5} \cite{PDT3.5} is a 2018 edition
of the core \textit{Prague Dependency Treebank}. The Prague Dependency
Treebank 3.5 contains the same texts as the previous versions since 2.0, and
is divided into \texttt{train}, \texttt{dtest}, and \texttt{etest} subparts,
where \texttt{dtest} is used as a development set and \texttt{etest} as a test
set. The dataset consists of several layers -- the morphological \texttt{m}-layer
is the largest and contains morphological annotations (POS
tags and lemmas), the analytical \texttt{a}-layer contains labeled dependency
trees, and the \texttt{t}-layer is the smallest and contains tectogrammatical
trees. The statistics of PDT 3.5 sizes is presented in
Table~\ref{table:PDT_3.5}.

A detailed description of the morphological system can be found in
\cite{hajic04}, a specification of the syntactic annotations
has been presented in~\cite{hajic98}. We note that in PDT, lemmas with the same word form are disambiguated using a number suffix -- for example, English lemmas for the word forms \texttt{can} (noun) and \texttt{can} (verb) would be annotated as \texttt{can-1} and \texttt{can-2}.

In evaluation, we compute:
\begin{citemize}
  \item POS tagging accuracy,
  \item lemmatization accuracy,
  \item unlabeled attachment score (UAS),
  \item labeled attachment score (LAS).
\end{citemize}

\begin{table}[t]
  \begin{center}
    \begin{tabular}{|l||rr|rr|}
      \hline
      \multirow{2}{*}{Part} & \multicolumn{2}{c|}{Morphological \texttt{m}-layer} & \multicolumn{2}{c|}{Analytical \texttt{a}-layer} \\\cline{2-5}
      & \multicolumn{1}{c}{Words} & \multicolumn{1}{c|}{Sentences} & \multicolumn{1}{c}{Words} & \multicolumn{1}{c|}{Sentences} \\\hline\hline
      Train       &~$1\,535\,826$ & $90\,828$ &~$1\,171\,190$ & $68\,495$ \\\hline
      Development & $201\,651$ & $11\,880$ & $158\,962$ & $9\,270$ \\\hline
      Test        & $219\,765$ & $13\,136$ & $173\,586$ & $10\,148$ \\\hline
    \end{tabular}
    \caption{Size of morphological and analytical annotations of PDT 3.5 train/development/test sets.}
    \label{table:PDT_3.5}
  \end{center}
\end{table}

\subsubsection{Universal Dependencies}

The \textit{Universal
Dependencies} project
\cite{UD} seeks to develop cross-linguistically consistent treebank annotation
of morphology and syntax for many languages. We evaluate the Czech PDT treebank of UD 2.3 \cite{UD2.3-data},
which is an automated conversion of PDT 3.5 \texttt{a}-layer to Universal
Dependencies annotation. The original POS tags are used to generate
\textbf{UPOS} (universal POS tags), \textbf{XPOS} (language-specific POS tags,
in this case the original PDT tags), and \textbf{Feats} (universal
morphological features). The UD lemmas are the raw textual lemmas, so the
discriminative numeric suffix of PDT is dropped. The dependency trees are
converted according to the UD guidelines, adapting both the unlabeled trees and the
dependency labels.

To compute the evaluation scores, we use the official \textit{CoNLL 2018 Shared Task:
Multilingual Parsing from Raw Text to Universal Dependencies} \cite{CoNLL2018}
evaluation script, which produces the following metrics:
\begin{citemize}
  \item \textbf{UPOS} -- universal POS tags accuracy,
  \item \textbf{XPOS} -- language-specific POS tags accuracy,
  \item \textbf{UFeats} -- universal subset of morphological features accuracy,
  \item \textbf{Lemmas} -- lemmatization accuracy,
  \item \textbf{UAS} -- unlabeled attachment score, \textbf{LAS} -- labeled attachment score,
  \item \textbf{MLAS} -- morphology-aware LAS, \textbf{BLEX} -- bi-lexical dependency score.
\end{citemize}

\subsubsection{Czech Named Entity Corpus}

The \textit{Czech Named Entity Corpus 1.1} \cite{Sevcikova2007,CNEC1.1-data} is a corpus of $5\,868$ Czech sentences with manually annotated $33\,662$ Czech named entities, classified according to a two-level hierarchy of $62$ named entities.

The \textit{Czech Named Entity Corpus 2.0} \cite{CNEC2.0-data} contains $8\,993$ Czech sentences with manually annotated $35\,220$ Czech named entities, classified according to a two-level hierarchy of $46$ named entities.

We evaluate the NER task with the official CNEC evaluation script. Similarly to previous literature
\cite{Sevcikova2007,Strakova2016} etc., the script only evaluates the first round annotation classes for the CNEC 1.1. For the CNEC 2.0, the script evaluates all annotated classes.

\section{Neural Architectures}

All our neural architectures are recurrent neural networks (RNNs). The POS
tagging, lemmatization and dependency parsing is performed with the
\textit{UDPipe 2.0} (Section~\ref{section:architectures-UDPipe2.0}) and NER is performed with our new sequence-to-sequence model (Section~\ref{section:architectures-NER}).

\subsection{POS Tagging, Lemmatization, and Dependency Parsing}
\label{section:architectures-UDPipe2.0}

\begin{figure*}[t]
  \begin{center}
    \includegraphics[width=\hsize]{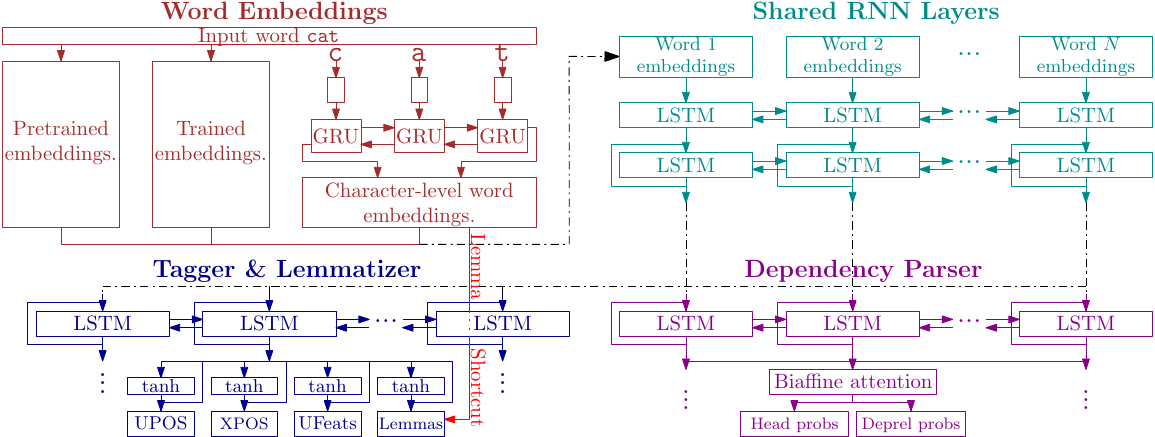}
  \end{center}
  \caption{UDPipe 2.0 architecture overview.}
  \label{figure:udpipe}
\end{figure*}

We perform POS tagging, lemmatization and dependency parsing using
\textit{UDPipe~2.0} \cite{UDPipe2.0}, one of the
three winning systems of the \textit{CoNLL 2018 Shared Task: Multilingual
Parsing from Raw Text to Universal Dependencies} \cite{CoNLL2018} and an
overall winner of \textit{The 2018 Shared Task on Extrinsic Parser Evaluation}
\cite{EPE2018}. An overview of this architecture is presented in
Figure~\ref{figure:udpipe} and the full details of the architecture
and the training procedure are available in~\cite{UDPipe2.0}.

\subsubsection{POS Tagging and Lemmatization}

The tagger employs a standard bi-LSTM architecture. After embedding input
words, three bidirectional LSTM~\cite{Hochreiter:1997:LSTM} layers are performed,
followed by a softmax output layers for POS tags and lemmas. While a classification output layer is natural for POS tags, we also apply it to lemmatization and generate lemmas by classifying the input words into lemma generation rules,
therefore considering lemmatization as another tagging task.

We construct a lemma generation rule from a given form and lemma as follows:
\begin{citemize}
  \item We start by finding the longest continuous substring of the form and
    the lemma. If it is empty, we use the lemma itself as the class.
  \item If there is a common substring of the form and the lemma, we compute
    the shortest edit script converting the prefix of the form into the prefix
    of the lemma, and the shortest edit script converting the suffix of the
    form to the suffix of the lemma. The edit scripts permit the
    operations {\small\verb|delete_current_char|} and
    {\small\verb|insert_char(c)|}.
  \item All above operations are performed case insensitively. To indicate
    correct casing of the lemma, we consider the lemma to be a concatenation of
    segments, where each segment is composed of either a sequence of lowercase
    characters, or a sequence of uppercase characters. We represent the lemma
    casing by encoding the beginning of every such segment, where the offsets
    in the first half of the lemma are computed relatively to the start of the
    lemma, and the offsets in the second half of the lemma are computed
    relatively to the end of the lemma.
\end{citemize}

\subsubsection{Dependency Parsing}

The dependency parsing is again predicted using \textit{UDPipe 2.0} architecture.
After embedding input words, three bidirectional
LSTM~\cite{Hochreiter:1997:LSTM} layers are again performed, followed
by a biaffine attention layer \cite{dozat:2016} producing labeled dependency
trees.

In our evaluation we do not utilize gold POS tags and lemmas on the test set
for dependency parsing. Instead, we consider three ways of employing them
during parsing:
\begin{citemize}
  \item not using them at all;
  \item adding predicted POS tags and lemmas on input;
  \item perform joint training of POS tags, lemmatization, and dependency
    parsing. In this case, we share first two bidirectional LSTM layers
    between the tagger and the parser.
\end{citemize}

\subsubsection{Input Embeddings}

In our \textbf{baseline} model, we use the end-to-end word embeddings and
also character-level word embeddings (bidirectional GRUs, \cite{Cho2014,Graves2005,Ling2015} of dimension $256$) trained specifically for the task.

Our architecture can optionally employ the following additional inputs

\begin{citemize}
  \item \textbf{pretrained word embeddings (WE):} For the PDT experiments, we generate the word embeddings with \texttt{word2vec}\footnote{With options \texttt{-size 300 -window 5 -negative 5 -iter 1 -cbow 0}.} on a concatenation of large raw Czech corpora\footnote{%
  The concatenated corpus has approximately 4G words, two thirds of them from SYN v3 \cite{synv3}.
  }
  available from the LINDAT/CLARIN repository.\footnote{\url{https://lindat.cz}} For UD Czech, we use FastText word embeddings \cite{FastText} of dimension $300$, which we pretrain on Czech Wikipedia using segmentation and tokenization trained from the UD data.\footnote{We use \texttt{-minCount 5 -epoch 10 -neg 10} options to generate the embeddings.}
  \item \textbf{BERT} \cite{BERT}: Pretrained contextual word embeddings of dimension $768$ from the \texttt{Base} model.\footnote{We use the BERT-Base Multilingual Uncased model from \url{https://github.com/google-research/bert}.} We average the last four layers of the BERT model to produce the embeddings. Because BERT utilizes word pieces, we decompose UD words into appropriate subwords and
    then average the generated embeddings over subwords belonging to the same word.
  \item \textbf{Flair} \cite{Akbik}: Pretrained contextual word embeddings of dimension $4096$.
\end{citemize}

\subsubsection{POS Tags and Lemmas Decoding}

Optionally, we employ a morphological dictionary MorfFlex~\cite{morfflex}
during decoding. If the morphological dictionary is used,
it may produce analyses for an input word as \textit{(POS tag, lemma)}
pairs. If any are generated, we choose the pair with maximum likelihood
given by both the POS tag and lemmatization model.

\subsection{Named Entity Recognition}
\label{section:architectures-NER}

We use a novel approach  \cite{NestedNer} for nested named entity recognition (NER) to capture
the nested entities in the Czech Named Entity Corpus.
The nested entities are encoded in a sequence and the problem of
nested NER is then viewed as a sequence-to-sequence (seq2seq) problem, in which
the input sequence consists of the input tokens (forms) and the output
sequence of the linearized entity labels.

The system is a encoder-decoder architecture. The encoder is a bi-directional
LSTM and the decoder is a LSTM. The encoded labels are predicted one by one by
the decoder, until the decoder outputs the \texttt{"<eow>"} (end of word) label
and moves to the next token. We use a hard attention on the word whose label(s)
is being predicted.

We train the network using the lazy variant of the Adam optimizer \cite{Adam},
which only updates accumulators for variables that appear in the current
batch,\footnote{\texttt{tf.contrib.opt.lazyadamoptimizer} from
\url{www.tensorflow.org}} with parameters $\beta_1=0.9$ and $\beta_2=0.98$. We
use mini-batches of size~$8$. As a regularization, we apply dropout with rate
$0.5$ and the word dropout replaces $20\%$ of words by the unknown token to
force the network to rely more on context. We did not perform any complex
hyperparameter search.

In this model, we use the following word- and character-level word embeddings:

\begin{citemize}
  \item \textbf{pretrained word embeddings:} We use the FastText \cite{FastText} word embeddings of dimension $300$ from the publicly available Czech model.\footnote{\url{https://fasttext.cc/docs/en/crawl-vectors.html}}
  \item \textbf{end-to-end word embeddings:} We embed the input forms and lemmas ($256$
    dimensions) and POS tags (one-hot).\footnote{POS tagging and lemmatization
    done with MorphoDiTa \cite{MorphoDiTa2014},
    \url{http://ufal.mff.cuni.cz/morphodita}.}
  \item \textbf{end-to-end character-level word embeddings:} We use
    bidirectional GRUs \cite{Cho2014,Graves2005} of dimension $128$ in line
    with \cite{Ling2015}: we represent every Unicode character with a vector of
    dimension $128$, and concatenate GRU outputs for forward and reversed word
    characters.
\end{citemize}

Optionally, we add the \textbf{BERT} \cite{BERT} and the \textbf{Flair}
\cite{Akbik} contextualized embeddings in the same way as in the UDPipe~2.0
(Section~\ref{section:architectures-UDPipe2.0}).

\section{Results}

\subsection{POS Tagging and Lemmatization on PDT 3.5}
\label{section:pdt_morpho}

\begin{table}[p]
  \begin{center}
    \catcode`! = 13\def!{\itshape}
    \catcode`@ = 13\def@{\bfseries}
    \footnotesize\renewcommand*{\arraystretch}{0.93}
    \begin{tabular}{|c|c|c||c|c|c||c|c|c|}
      \hline
      \multirow{2}{*}{~~~~WE~~~~} & \multirow{2}{*}{~~BERT~~} & \multirow{2}{*}{~~~Flair~~~} & \multicolumn{3}{c||}{Without Dictionary} & \multicolumn{3}{c|}{With Dictionary} \\
      \cline{4-9}
      &&& POS Tags & Lemmas & Both & POS Tags & Lemmas & Both \\
      \hline
      \NO & \NO & \NO &     96.88\% &  98.35\% &  96.21\% &  97.31\% &  98.80\% &  96.89\% \\
      \YES & \NO & \NO &    97.43\% &  98.55\% &  96.77\% &  97.59\% &  98.82\% &  97.18\% \\
      \NO & \NO & \YES &    97.24\% &  98.49\% &  96.61\% &  97.54\% &  98.86\% &  97.14\% \\
      \YES & \NO & \YES &   97.53\% &  98.63\% &  96.91\% &  97.69\% &  98.88\% &  97.28\% \\
      \NO & \YES & \NO &    97.67\% &  98.63\% &  97.02\% &  97.91\% &  98.94\% &  97.51\% \\
      \YES & \YES & \NO &   97.86\% &  98.69\% &  97.21\% &  98.00\% &  98.96\% &  97.59\% \\
      \NO & \YES & \YES &   97.80\% &  98.67\% &  97.16\% &  98.00\% &  98.96\% &  97.59\% \\
      \YES & \YES & \YES & @97.94\% & @98.75\% & @97.31\% & @98.05\% & @98.98\% & @97.65\% \\
      \hline\hline
      \multicolumn{3}{|l||}{!Mor\v{c}e (2009) \cite{spoustova09}}& !---     & !---     & !--- & !95.67\%\rlap{$^\dagger$} & !---     & !--- \\
      \multicolumn{3}{|l||}{!MorphoDiTa (2016) \cite{ACL2014}}   & !---     & !---     & !--- & !95.55\% & !97.85\% & !95.06\% \\
      \multicolumn{3}{|l||}{!LemmaTag (2018) \cite{EMNLP2018}}   & !96.90\% & !98.37\% & !--- & !---     & !---     & !--- \\
      \hline
    \end{tabular}
    \caption{POS tagging and lemmatization results (accuracy) on PDT 3.5.
    {\bfseries Bold} indicates the best result, {\itshape italics} related
    work. $^\dagger$Reported on PDT 2.0, which has the same underlying corpus,
    with minor changes in morphological annotation (our model results differ
    at $0.1\%$ on PDT 2.0).
    }
    \label{table:POS_PDT}
  \end{center}
  
  \begin{center}
    \catcode`! = 13\def!{\itshape}
    \catcode`@ = 13\def@{\bfseries}
    \footnotesize\renewcommand*{\arraystretch}{0.93}
    \begin{tabular}{|c|c|c||c|c||c|c|}
      \hline
      POS Tags, & \multirow{2}{*}{~BERT~} & \multirow{2}{*}{Flair} & UAS (unlabeled & LAS (labeled & \multirow{2}{*}{POS Tags} & \multirow{2}{*}{Lemmas} \\
      Lemmas &&& attachment score) & attachment score) & & \\\hline
      \NO & \NO & \NO            &  91.16\% & 87.35\% & --- & --- \\
      \NO & \NO & \YES           &  91.38\% & 87.69\% & --- & --- \\
      \NO & \YES & \NO           &  92.75\% & 89.46\% & --- & --- \\
      \NO & \YES & \YES          &  92.76\% & 89.47\% & --- & --- \\
      Predicted on input & \YES & \YES & 92.84\% & 89.62\% & --- & --- \\
      Joint prediction & \NO & \NO    &  91.69\% & 88.16\% & !97.33\% & !98.42\% \\
      Joint prediction & \NO & \YES   &  91.89\% & 88.42\% & !97.48\% & !98.42\% \\
      Joint prediction & \YES & \NO   &  93.01\% & 89.74\% & !97.62\% & !98.49\% \\
      Joint prediction & \YES & \YES  & @93.07\% &@89.89\% & !97.72\% & !98.51\% \\
      \hline
      !Gold on input   & \YES & \YES  & !92.95\% &!89.89\% & --- & --- \\
      \hline
      \multicolumn{3}{|c||}{!POS tagger trained on 3.5 \texttt{a}-layer} & --- & --- & !97.82\% & !98.66\% \\
      \hline
    \end{tabular}
    \caption{Dependency tree parsing results on PDT 3.5 \texttt{a}-layer.
    {\bfseries Bold} indicates the best result, {\itshape italics} POS tagging and
    lemmatization results. For comparison, we report results of a parser
    trained using gold POS tags and lemmas, and of a~tagger trained
    on \texttt{a}-layer (both also in {\itshape italics}).}
    \label{table:Parsing_PDT_Ablations}
  \end{center}
  
    \begin{center}
    \catcode`! = 13\def!{\itshape}
    \catcode`@ = 13\def@{\bfseries}
    \footnotesize\renewcommand*{\arraystretch}{0.93}
    \begin{tabular}{|l||c|c|}
      \hline
      \multicolumn{1}{|c||}{\multirow{2}{*}{System}} & UAS (unlabeled & LAS (labeled \\
       & attachment score) & attachment score) \\\hline
       Our best system (joint prediction, BERT, Flair) & @93.10\% & @89.93\% \\\hline
       !Holan and \v{Z}abokrtsk\'{y} (2006) \cite{holan:zabokrtsky:2006} & !85.84\% & --- \\
       !Nov\'{a}k and \v{Z}abokrtsk\'{y} (2007) \cite{novak:zabokrtsky:2007} & !84.69\% & --- \\
       !Koo et al. (2010) \cite{koo:2010}$^\dagger$& !87.32\% & --- \\
       !Treex framework (using MST parser\&manual rules) \cite{treex}$^\ddagger$ & !83.93\% & !77.04\%\\
      \hline\hline
      \multicolumn{3}{|c|}{PDT 2.0 subset in CoNLL 2007 shared task; manually annotated POS tags available.} \\
      \hline
       !Nakagawa (2007) \cite{nakagawa:2007} & !86.28\% & !80.19\% \\
      \hline\hline
      \multicolumn{3}{|c|}{PDT 2.0 subset in CoNLL 2009 shared task; manually annotated POS tags available.} \\
      \hline
       !Gesmundo et al. (2009) \cite{gesmundo:2009} & --- & !80.38\% \\
      \hline
    \end{tabular}
    \caption{Dependency tree parsing results on PDT 2.0 \texttt{a}-layer.
    {\bfseries Bold} indicates the best result, {\itshape italics} related work.
    $^\dagger$Possibly using gold POS tags. $^\ddagger$Results as of 23 Mar
    2019.}
    \label{table:Parsing_PDT_SOTA}
  \end{center}
\end{table}

The POS tagging and lemmatization results are presented in Table~\ref{table:POS_PDT}.
The word2vec word embeddings (WE) considerably increase performance compared to the baseline,
especially in POS tagging. When only Flair embeddings are added to the baseline, we also observe
an improvement, but not as high. We hypothesise that the lower performance
(in contrast with the results reported in \cite{Akbik}) is caused by the size of the
training data, because we train the word2vec WE on considerably larger
dataset than the Czech Flair model. However, when WE and Flair embeddings are combined, performance moderately
increases, demonstrating that the two embedding methods produce at least partially
complementary representations.

The BERT embeddings alone bring highest improvement in performance. Furthermore,
combination with WE or Flair again yields performance increase. The best results
are achieved by exploiting all three embedding methods, substantially exceeding
state-of-the-art results.

Utilization of morphological dictionary improves prediction accuracy. However, as
the performance of a model itself increases, the gains obtained by the morphological dictionary
diminishes -- for a model without any pretrained embeddings, morphological dictionary
improves POS tagging by and lemmatization by $0.43\%$ and $0.45\%$, while
the best performing model gains only $0.11\%$ and $0.23\%$.

\subsection{Dependency Parsing on PDT 3.5}
\label{section:pdt_syntax}

The evaluation of the contextualized embeddings methods as well as various ways of POS tag
utilization is presented in Table~\ref{table:Parsing_PDT_Ablations}. Without POS tags and lemmas, the Flair embeddings bring only a slight improvement in dependency parsing
when added to WE. In contrast, BERT embeddings employment results in substantial gains,
increasing UAS and LAS by 1.6\% and 2.1\%. A combination of BERT and Flair embeddings does not result
in any performance improvement, demonstrating that BERT syntactic representations encompass the Flair embeddings.

When introducing POS tags and lemmas predicted by the best model from Section~\ref{section:pdt_morpho}
as inputs for dependency parsing, the performance increases only slightly.
A better way of POS tags and lemmas exploitation is achieved in a joint model, which predicts
POS tags, lemmas, and dependency trees simultaneously. Again, BERT embeddings bring significant
improvements, but in contrast to syntax parsing only, adding Flair embeddings to BERT results
in moderate gain -- we hypothesise that the increase is due to the complementary morphological
information present in Flair embeddings (cf. Section~\ref{section:pdt_morpho}).
Note that the joint model achieves better parsing accuracy than the one given gold POS tags
and lemmas on input. However, the POS tags and lemmas predicted by the joint model are of slightly
lower quality compared to a standalone tagger of the best configuration from Section~\ref{section:pdt_morpho}.

Table~\ref{table:Parsing_PDT_SOTA} compares our best model with state-of-the-art results on PDT 2.0
(note that some of the related work used only a subset of PDT 2.0 and/or utilized gold morphological
annotation). To our best knowledge, research on PDT parsing was performed mostly in
the first decade of this century, therefore even our baseline model substantially surpasses
previous works. Our best model with contextualized embeddings achieves nearly 50\% error
reduction both in UAS and LAS.

\subsection{POS Tagging, Lemmatization and Dependency Parsing on Universal
Dependencies}

\begin{table}[t]
  \begin{center}
    \catcode`@ = 13\def@{\bfseries}
    \catcode`! = 13\def!{\itshape}
    \footnotesize\renewcommand*{\arraystretch}{0.93}
    \begin{tabular}{|c|c|c||c|c|c|c|c|c|c|c|}
      \hline
      ~~~WE~~~ & ~BERT~ & ~~Flair~~ & UPOS  & XPOS & UFeats & Lemmas & UAS & LAS & MLAS & BLEX \\\hline
      \NO  & \NO & \NO   & 99.06 & 96.73 & 96.69 & 98.80 & 92.93 & 90.75 & 84.99 & 87.68\\\hline
      \YES & \NO & \NO   & 99.18 & 97.28 & 97.23 & 99.02 & 93.33 & 91.31 & 86.15 & 88.60\\\hline
      \NO  & \NO & \YES  & 99.16 & 97.17 & 97.13 & 98.93 & 93.33 & 91.33 & 86.19 & 88.56\\\hline
      \YES & \NO & \YES  & 99.22 & 97.41 & 97.36 & 99.07 & 93.48 & 91.49 & 86.62 & 88.89\\\hline
      \NO  & \YES & \NO  & 99.25 & 97.46 & 97.41 & 99.00 & 94.26 & 92.34 & 87.53 & 89.79\\\hline
      \YES & \YES & \NO  & 99.31 & 97.61 & 97.55 & 99.06 & 94.27 & 92.34 & 87.75 & 89.91\\\hline
      \YES & \YES & \YES &@99.34 &@97.71 &@97.67 &@99.12 &@94.43 &@92.56 &@88.09 &@90.22\\\hline
      \hline
      \multicolumn{11}{|c|}{CoNLL 2018 Shared Task results on Czech PDT UD 2.2 treebank,} \\
      \multicolumn{11}{|c|}{from raw text (without gold segmentation and tokenization).} \\\hline
      \multicolumn{3}{|l|}{Our best system} & @99.24 & @97.63 & @97.62 & @99.08 & @93.69 &@91.82 & @87.57 & @89.60\\\hline
      \multicolumn{3}{|l|}{!HIT-SCIR (2018)~\cite{HIT-SCIR}} & !99.05 & !96.92 & !92.40 & !97.78 & !93.44 & !91.68 & !80.57 & !87.91\\\hline
      \multicolumn{3}{|l|}{!TurkuNLP (2018)~\cite{TurkuNLP}} & !98.74 & !95.44 & !95.22 & !98.50 & !92.57 & !90.57 & !83.16 & !87.63\\\hline
    \end{tabular}
    \caption{Czech PDT UD 2.3 results for POS tagging (UPOS: universal POS, XPOS:
    language-specific POS, UFeats: universal morphological features),
    lemmatization and dependency parsing (UAS, LAS, MLAS, and BLEX scores).
    {\bfseries Bold} indicates the best result, {\itshape italics} related
    work.}
    \label{table:UD}
  \end{center}
  
  \begin{center}
    \catcode`@ = 13\def@{\bfseries}
    \catcode`! = 13\def!{\itshape}
    \footnotesize\renewcommand*{\arraystretch}{0.93}
    \begin{tabular}{|c|c||cc|cc|}
      \hline
      \multirow{2}{*}{\kern1.4em BERT\kern1.4em} & \multirow{2}{*}{\kern1.6em Flair\kern1.6em} & \multicolumn{2}{c|}{CNEC 1.1} & \multicolumn{2}{c|}{CNEC 2.0} \\
      \cline{3-6}
      && \multicolumn{1}{c}{\kern1em Types\kern1em} & \multicolumn{1}{c|}{Supertypes} & \multicolumn{1}{c}{\kern1em Types\kern1em} & \multicolumn{1}{c|}{Supertypes} \\
      \hline
      \NO & \NO    & 82.96 & 86.80   & \sout{80.47}  79.04~~ & \sout{85.15} 83.90 \\\hline
      \NO & \YES   & 83.55 & 87.62   & \sout{81.65}  80.30~~ & \sout{85.96} 84.72 \\\hline
      \YES & \NO   & 86.73 & 89.85   & \sout{86.23} @84.66~~ & \sout{89.37} @88.02 \\\hline
      \YES & \YES  & @86.88 & @89.91 & \sout{85.52}  84.27~~ & \sout{89.01} 87.81 \\\hline
      \hline
      \multicolumn{2}{|l||}{!Konkol et al. (2013)     \cite{konkol13crf}}  & --     & !79.00  & --    & -- \\\hline
      \multicolumn{2}{|l||}{!Strakov\'a et al. (2013) \cite{strakova13}}   & !79.23  & !82.82  & --    & -- \\\hline
      \multicolumn{2}{|l||}{!Strakov\'a et al. (2016) \cite{Strakova2016}} & !81.20  & !84.68 & !79.23 & !82.78 \\\hline
    \end{tabular}
    \caption{Named entity recognition results (F1) on the Czech Named Entity
    Corpus. {\bfseries Bold} indicates the best result, \textit{italics}
    related work. \bfseries EDIT 12 Apr 2021: The CNEC 2.0 results were
    incorrectly evaluated; we now provide the correct results.}
    \label{table:NER}
  \end{center}
\end{table}

Table~\ref{table:UD} shows the performance of analyzed embedding methods in a joint model performing
POS tagging, lemmatization, and dependency parsing on Czech PDT UD 2.3 treebank. This treebank is derived from PDT 3.5 \texttt{a}-layer,
with original POS tags kept in XPOS, and the dependency trees and lemmas modified
according to UD guidelines.

We observe that the word2vec WEs perform similarly to Flair embeddings in this setting.
Our hypothesis is that the word2vec WEs performance loss (compared to WEs in Section~\ref{section:pdt_morpho}) is caused by using a considerably smaller raw corpus to pretrain the WEs (Czech Wikipedia with 785M words, compared to 4G words used in Section~\ref{section:pdt_morpho}), due to licensing reasons. BERT embeddings once more deliver the highest improvement, especially in dependency parsing, and our best model employs all three embedding methods.

In the previous ablation experiments, we used the gold segmentation and
tokenization in the Czech PDT UD 2.3 treebank. For comparison with state of the
art, Czech PDT UD 2.2 treebank without gold segmentation and tokenization is
used in evaluation, according to the CoNLL 2018 shared task training and
evaluation protocol. Our system reuses segmentation and tokenization produced
by UDPipe 2.0 in the CoNLL 2018 shared task and surpasses previous works substantially in all
metrics (bottom part of Table~\ref{table:UD}).

Comparing the results with a joint tagging and parsing PDT 3.5 model from Table~\ref{table:PDT_3.5},
we observe that the XPOS results are nearly identical as expected. Lemmatization on the UD treebank
is performed without the discriminative numeric suffixes (see Section~\ref{section:datasets})
and therefore reaches better performance. Both UAS and LAS are also better on the UD treebank, which
we assume is caused by the different annotation scheme.

\subsection{Named Entity Recognition}

Table~\ref{table:NER} shows NER results (F1 score) on CNEC 1.1 and CNEC 2.0.
Our sequence-to-sequence (seq2seq) model which captures the nested entities,
clearly surpasses the current Czech NER state of the art. Furthermore,
significant improvement is gained when adding the contextualized word
embeddings (BERT and Flair) as optional input to the LSTM encoder. The strongest model is a combination of the sequence-to-sequence architecture with both BERT and Flair contextual word embeddings.

\section{Conclusion}

We have presented an evaluation of two contextualized embeddings methods, namely BERT and Flair.
By utilizing these embeddings as input to deep neural networks, we have achieved state-of-the-art
results in several Czech text processing tasks, namely in POS tagging, lemmatization, dependency
parsing and named entity recognition.

\section*{Acknowledgements}

The work described herein has been supported by OP VVV VI LINDAT/CLARIN project
(CZ.02.1.01/0.0/0.0/16\_013/0001781) and it has been supported and has been
using language resources developed by the LINDAT/CLARIN project (LM2015071) of
the Ministry of Education, Youth and Sports of the Czech Republic.

\bibliographystyle{splncs03}
\bibliography{tsd2019}

\end{document}